\documentclass[sigconf]{acmart}

\usepackage{booktabs} 
\usepackage[normalem]{ulem}
\useunder{\uline}{\ul}{}

\usepackage{float}

\usepackage[utf8]{inputenc}
\usepackage[english]{babel}

\usepackage{bm}

\usepackage{graphicx}

\usepackage{amsmath}
\usepackage{amsfonts}
\usepackage{amssymb}

\setcopyright{none}

\acmDOI{}

\acmISBN{}

\acmConference[DSJM2018]{2018 KDD Data Science, Journalism and Media }{August 2018}{London, UK}
\acmYear{2018}
\copyrightyear{2018}

\begin{document}
\title{Content-driven, unsupervised clustering of news articles through multiscale graph partitioning}

\author{M.\ Tarik Altuncu}
\orcid{0000-0003-0516-1201}
\affiliation{%
  \institution{Department of Mathematics\newline 
  Imperial College London}
  \streetaddress{South Kensington Campus}
  \state{United Kingdom}
 \postcode{SW7 2AZ}
}
\email{m.altuncu16@imperial.ac.uk}

\author{Sophia N.\ Yaliraki}
\affiliation{%
  \institution{Department of Chemistry\newline Imperial College London}
  \streetaddress{South Kensington Campus}
  \state{United Kingdom}
  \postcode{SW7 2AZ}
}
\email{s.yaliraki@imperial.ac.uk}

\author{Mauricio Barahona}
\orcid{0000-0002-1089-5675}
\affiliation{%
  \institution{Department of Mathematics\newline Imperial College London}
  \streetaddress{South Kensington Campus}
  \state{United Kingdom}
  \postcode{SW7 2AZ}
}
\email{m.barahona@imperial.ac.uk}

\renewcommand{\shortauthors}{M.T. Altuncu et al.}

\begin{abstract}
The explosion in the amount of news and journalistic content being generated across the globe, coupled with extended and instantaneous access to information through online media, makes it difficult and time-consuming to monitor news developments and opinion formation in real time. There is an increasing need for tools that can pre-process, analyse and classify raw text to  extract interpretable content; specifically, identifying topics and content-driven groupings of articles. 
We present here such a methodology that brings together powerful vector embeddings from Natural Language Processing with tools from Graph Theory that exploit diffusive dynamics on graphs to reveal natural partitions across scales. 
Our framework uses a recent deep neural network text analysis methodology (Doc2vec) to represent text in vector form and then applies a multi-scale community detection method (Markov Stability) to partition a similarity graph of document vectors. 
The method allows us to obtain clusters of documents with similar content, at different levels of resolution, in an unsupervised manner. 
We showcase our approach with the analysis of a corpus of 9,000 news articles published by Vox Media over one year. 
Our results show consistent groupings of documents according to content without \textit{a priori} assumptions about the number or type of clusters to be found. 
The multilevel clustering reveals a quasi-hierarchy of topics and subtopics with increased intelligibility and improved topic coherence as compared to external taxonomy services and standard topic detection methods. 
\end{abstract}

%
%



\maketitle

\section{introduction}

The production of news content is growing at an astonishing rate. To help manage and monitor such sheer amount of text, there is an increasing need to develop efficient methods that can provide insight into emerging content areas, and help with the stratification of articles and written pieces according to `topics' that follow \emph{intrinsically} from content similarity. This is in contrast with traditional approaches to article classification, typically based on keywords, word frequencies, and man-made hierarchies.
Methodologies that provide automatic clustering of articles based on content, directly from free text without external labels or categories, could thus provide alternative ways to monitor the generation and emergence of news content beyond standard, broad classes.

  The area of text mining and natural language processing (NLP) has a long history. Classic methods for topic detection have been usually based on characterisation of documents using weighted word frequencies (Bag-of-Words (BoW) representation) followed by statistical methods, notably Latent Dirichlet Allocation (LDA), to cluster documents into topics. However, such methods do not fully capture the highly meaningful chains of association present in natural language, nor the hierarchical relationships between topics at different levels of resolution.
Recently, powerful methods based on deep neural networks have been introduced to represent words and documents through vector embeddings~\cite{doc2vec} and shown marked improvements over statistical BoW descriptions. 
Some approaches have considered clustering based on such vectors~\citep{Hashimoto2016TopicReviews} but those approaches lack a full multiscale graph analysis leading to a hierarchy of topics. Network-theoretic methods applied to text analysis~\citep{PhysRevX.5.011007} suffer from the same limitation as well as lacking the additional power afforded by text vector embeddings.

Here we present a method that combines the advantages of both: paragraph vectors to represent text~\cite{doc2vec}, and a multi-scale community detection algorithm~\cite{pnasStability} applied to the similarity graph between document vectors. This approach allows us to obtain different graph partitions (from finer to coarser) corresponding to groupings of articles with consistent content at different levels of resolution (from more specific to more generic). Therefore the groupings of articles emerge directly in an unsupervised manner from the vector embedding, which captures both syntactic and semantic characteristics of the text, instead of fitting to pre-designed classifications. 

\section{methodology}

\subsection{Data}
As part of the DS+J 2017 Workshop~\citep{dsjm2017}, Vox Media made available a dataset containing all the articles published by Vox before March 21, 2017. \footnote{The corpus of Vox articles is accessible on \url{https://data.world/elenadata/vox-articles}.} 
We used this dataset and analysed the published content for the whole year of 2016. This corresponds to a corpus of 9021 articles in a wide range of topics, from politics to sport to human interest with a clear focus on the US, where Vox is based. Given that 2016 was an electoral year in the US, news about the election and the primaries feature heavily in the corpus, as do other events such as Brexit and the Olympic Games.

\subsection{Computational framework for the clustering}

\begin{figure}[t!]
\includegraphics[width=.99\linewidth,angle=0]{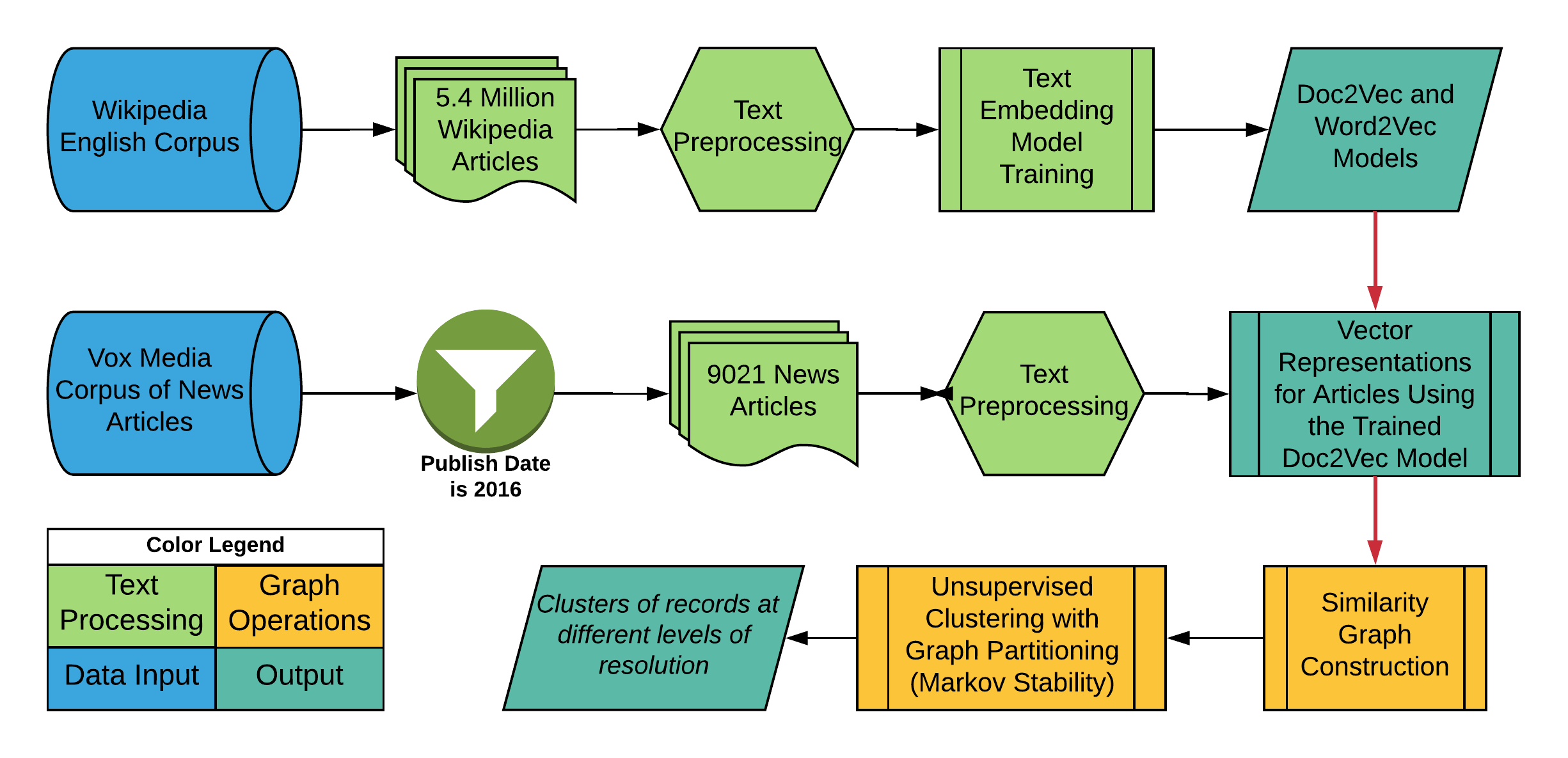}
\caption{
Schematic of the data pipeline for the analysis of the Vox Media corpus documents.
}
\label{fig:pipeline}
\end{figure}

A schematic of our computational pipeline is presented in Figure~\ref{fig:pipeline}. Our methodology consists of text processing and graph operations.

\subsubsection{\textbf{Text operations.}}

Before starting the analysis, it is necessary to apply standard pre-processing to the text. The pre-processing step tokenises documents into words, removing tokens with no distinct meaning (such as digits and stop words), and normalises the rest of the tokens to clean suffixes using stemming methods, as implemented in the NLTK module~\citep{nltk}. This protocol is applied to all the text we analyse.

It was also noticed that the raw articles contained repeated meaningless wrapper sentences (mostly header or footer scripts, directions to interact with multi-media content, signatures, etc), which introduce a considerable amount of noise in the analysis. To eliminate such scripted sentences in the news articles, we created a hashed dictionary with sentence tokens as keys and frequency of the sentence in the corpus as values. We then reconstructed the articles only with the sentences that have frequency up to 2. (Allowing for sentences with frequency of 2 capture the possibility that quotations from a press release might have been copied in multiple places of an original article.) This procedure reduces substantially (but not totally) the amount unwanted, meaningless wrapper phrases.

Our text operations continue with the training of a Doc2vec model~\cite{doc2vec} based on a corpus of 5.4 million articles from WikiPedia. Doc2vec, also known as paragraph vectors (PV), is a neural network model. It starts with the random initial vectors for each word and document and adjusts the vectors over iterations of the algorithm. In this case, we apply the PV-DBOW method which uses document vectors to predict a set of words sampled from the document. 
Each iteration covers this learning phase throughout the corpus of documents. Here, we use the Gensim~\citep{gensim} module with parameters as suggested in~\citep{jhlau}. In our final optimised model, the parameters are set to: {training method = dbow, number of dimensions for feature vectors size = 300, number of epochs = 10, window size = 5, minimum count = 20, number of negative samples = 5, random down-sampling threshold for frequent words = 0.001.}
By using a large Wikipedia corpus, our Doc2vec model is trained on a standard collection of English language documents on a variety of topics thus providing a good representation of the general text encountered in news articles. 

Once the Doc2vec model is trained on the Wikipedia corpus, we apply it to the Vox news articles (pre-processed and cleaned from wrappers). The model is used to infer a 300-dimensional paragraph vector for each of the 9021 articles. This paragraph vector, which encapsulates a variety of semantic and syntactic characteristics of the text, is used as the article descriptor in our analysis.

\subsubsection{}{\textbf{Graph operations.}}
Our aim is to find groups of articles with high in-group similarity and low out-of-group similarity based on paragraph vectors. This classic clustering problem can be approached in a myriad of ways, from simple hierarchical clustering to $k$-means to spectral clustering~\citep{spectral_clustering,Schaub2017}. We discuss the comparison to some of these standard methods in Section~\ref{sec:quality}. Here we take a graph-theoretical approach that uses multi-scale community detection to find groups in a similarity graph of the articles.

We start by obtaining the pairwise cosine similarity for all pairs of articles. 
These distances form a similarity matrix, which can be thought of as a weighted adjacency matrix of a complete graph. 
To construct a sparser graph that preserves the local geometry of the data and retains global connectivity in the dataset, we construct a geometric graph using the MST-kNN method with $k=13$~\citep{mstknn}. 
The result is a similarity graph $\mathcal{G}_S$ where the nodes are articles and the weighted edges represent similarities between them.

We apply Markov Stability~\footnote{The code for Markov Stability is open and accessible at \url{https://github.com/michaelschaub/PartitionStability} and \url{http://wwwf.imperial.ac.uk/~mpbara/Partition_Stability/}, last accessed on March 24, 2018}~\cite{pnasStability, Arxiv_Lambiotte_2008, Delvenne2013, Schaub2012ZoomingLens, LambiotteMarkovProcess} to the similarity graph $\mathcal{G}_S$ in order to extract the multi-scale community structure intrinsic to the graph. Markov Stability scans across all levels of resolution and finds consistent and robust partitions without imposing \textit{a priori} the number of communities to be found.

We summarise the Markov Stability (MS) graph partitioning algorithm here. For details, see~\cite{ pnasStability, Delvenne2013, Schaub2012ZoomingLens, LambiotteMarkovProcess}. 
Given the (symmetric) adjacency matrix $A_{NxN}$ of the (undirected) similarity graph $\mathcal{G}_S$, obtained as above, we define $k=A1$, the vector of degrees, and $D=diag(k)$. The random walk Laplacian matrix is defined as $\mathcal{L}=I_N-D^{-1}A$ where $I_N$ is the identity matrix of size $N$. The transition matrix of the associated continuous time Markov process taking place on the graph is given by $P(t)=e^{-t \mathcal{L}}$ for $t>0$~\citep{pnasStability,Arxiv_Lambiotte_2008, Delvenne2013, LambiotteMarkovProcess}. The time parameter in this process $t$ is denoted the Markov time.

MS is a hard clustering method that searches for partitions at each Markov time through the optimisation of a cost function, as follows. For a given partition, we have a 0-1 membership matrix $H$ that maps $N$ nodes to $C$ clusters (or communities). We then define the clustered autocovariance matrix:
\begin{eqnarray}
R(t,H) &=& H^T[\Pi P(t)-\pi^T\pi]H, \label{eq:diffusion} 
\end{eqnarray}
where $\pi$ is the steady-state distribution of the Markov process and $\Pi=\text{diag}(\pi)$. 
$R(t,H)$ is a $C \times C$ matrix, specific to the partition $H$ on the graph $\mathcal{G}_S$, and its element $R_{ij}(t,H)$ quantifies the probability that a random walker starting from community $i$ will end in community $j$ at time $t$ minus the probability that such event occurs by chance at stationarity.

With these definitions, we introduce the cost function for the graph partitioning optimisation. The Markov Stability of a partition with membership matrix $H$ at time $t$ is defined as the trace of the clustered autocovariance matrix of the diffusion process~\eqref{eq:MS}:
\begin{eqnarray}
r(t,H) &=& \text{trace} \left[R(t,H)\right]. \label{eq:MS}
\end{eqnarray}
From this definition, a partition $H$ that maximises $r(t,H)$ is one in which the diagonal elements of $R(t,H)$ are large and the off-diagonal elements small, i.e., such a partition is comprised of well-defined communities, in the sense that they \textit{retain within them} the probabilistic flow of the Markov process over time $t$.
Our aim is therefore to find partitions that maximise the Markov Stability, parametrically as a function of the Markov time $t$. 
The maximisation of~\eqref{eq:MS} is an NP-Hard problem (hence no guaranteed global optimality). However, there are efficient optimisation methods that work well in practice. Our implementation uses the Louvain Algorithm~\citep{louvain} which is both computationally efficient and known to give good results on benchmarks~\citep{LambiotteMarkovProcess,Arxiv_Lambiotte_2008,bacik_celegans}. 

\begin{figure}[h!]
\includegraphics[width=\linewidth] {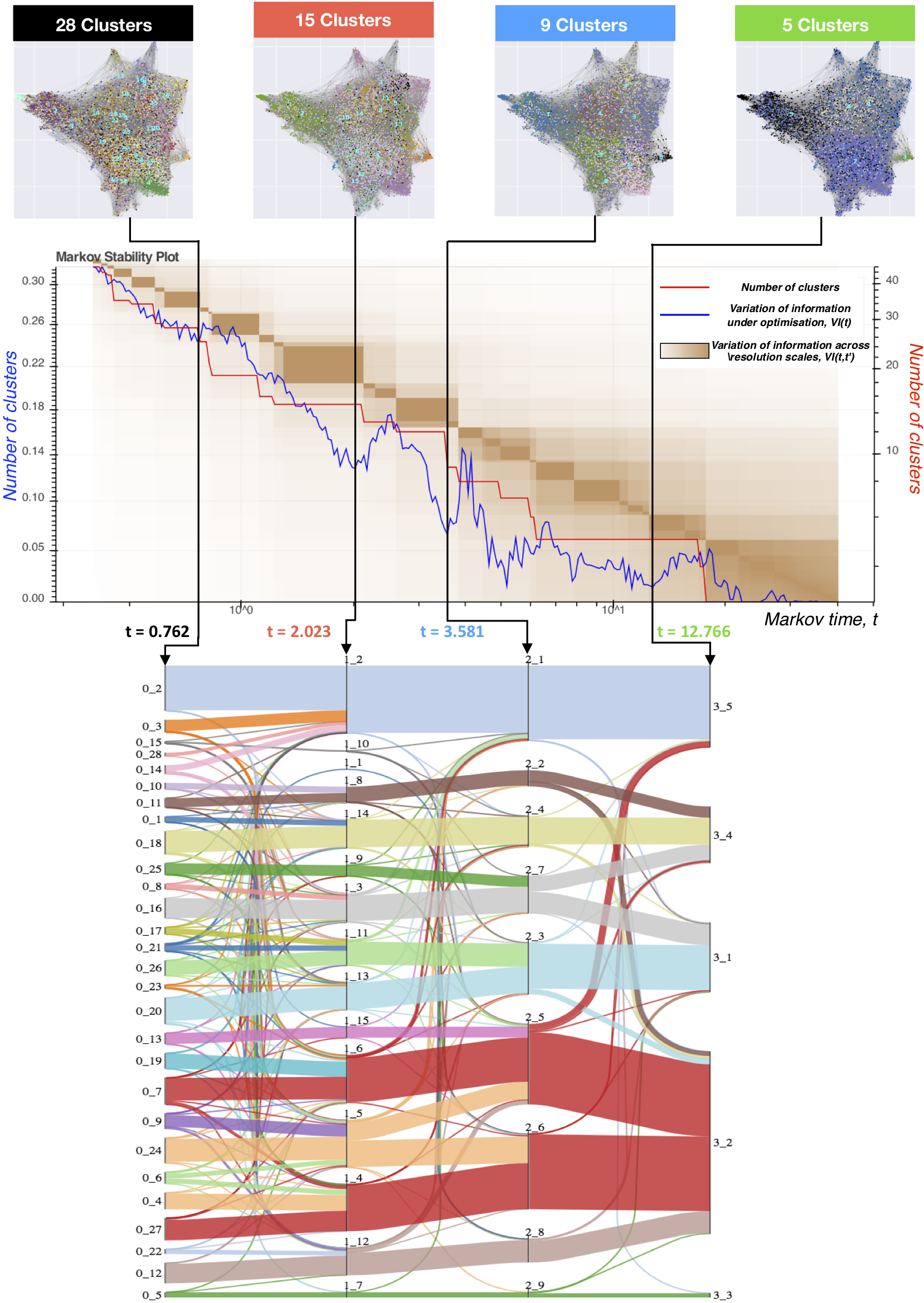}
\caption{
The top plot presents the results of the Markov Stability algorithm, showing the number of clusters of the optimized partition (red), the variation of Information $VI(t)$ for the ensemble of optimized solutions at each time (blue), and the variation of Information $VI(t,t')$ between the optimized partitions across Markov time (background colourmap). Relevant partitions are indicated by dips of $VI(t)$ and extended plateaux of $VI(t,t')$. We choose 4 levels with different resolutions (from 28 communities to 5) in our analysis. The Sankey diagram (bottom) illustrates how the communities of documents (indicated by numbers and colours) map across Markov time scales. Note that the community structure across scales presents a strong quasi-hierarchical character---a result of the analysis and the properties of the data as it is not imposed \textit{a priori}. The partitions for the four chosen levels ($t=0.762, 2.023, 3.581, 12.766$) are also shown as colourings on the similarity graph (MST-kNN with $k=13$).
}
\label{fig:MS}
\end{figure}

The dynamic sweeping provided by the Markov process is key to the MS community detection procedure. As $t$ grows, the Markov process explores larger horizons within the graph, and the communities become coarser. Hence the Markov time $t$ can be understood as providing a varying level of resolution. In other words, scanning $t$ provides a means of finding intrinsically good partitions of the graph at all time scales.
Indeed, not all the optimised partitions found in our scan are equally relevant. We look for robust and persistent partitions, as follows.  At each $t$, we run the Louvain algorithm 500 times with different initialisations and collect all the optimised partitions.  We then compute the mean Variation of Information~\citep{Meila2007} of this ensemble, $VI(t)$, which provides a measure of the difference between all the partitions found by running the algorithm repeatedly. A low value of $VI(t)$ indicates a highly reproducible result under the optimization, i.e., a robust partition.
In addition, we look for partitions that are persistent across Markov times (i.e., partitions that are optimal over an extended scale). To quantify this, we compute the variation of information between optimised partitions across time, $VI(t,t')$, and look for partitions found over prolonged Markov time spans.
Relevant partitions are those where $VI(t)$ shows a dip and $VI(t,t')$ has an extended plateau, thus indicating robustness and persistence~\citep{bacik_celegans,LambiotteMarkovProcess}.

We study robust MS partitions of the similarity graph of documents $\mathcal{G}_S$ obtained at selected resolution scales, from finer to coarser. The communities of nodes in a partition correspond to groups of documents with similar content. An example of this analysis is presented in Fig.~\ref{fig:MS}.

\subsection{\textit{A posteriori} analysis of clustering results}

\subsubsection{\textbf{Visualising membership with Sankey diagrammes.}}
The relationship between different partitions (found across scales or using different methods) is captured through Sankey diagrammes. This standard visualisation represents membership across groupings, and reveals visually the quasi-hierarchical structure in the data (see Figs.~\ref{fig:MS} and \ref{fig:15to5}) or the consistency of groups across partitions (see Fig.~\ref{fig:Calais_Google}).

\subsubsection{\textbf{Content summaries for clusters.}}
To aid our interpretation of the clusters of documents found with MS, we extract descriptive features \textit{a posteriori}. Specifically, for each cluster we generate word clouds to represent common words in it, and the paragraph vector for the cluster centroid, from which we obtain the nearest Wikipedia article, Vox Media article, and Word2vec words to each cluster (see Fig.~\ref{fig:15to5}).

\subsubsection{\textbf{Evaluating the quality of the clusterings.}}

The absence of a ground truth in our study leads us to use two different types of measures to quantify the consistency and relevance of content within the clusters of documents. We used these two measures to evaluate the quality of our clusters in comparison with other methods and against commercial news classification services.

\paragraph*{Measuring intrinsic Topic Coherence}\label{sec:tc}

To check for topic coherence within each cluster, we utilized the pointwise mutual information (PMI), an information-theoretical score that reflects semantic similarities between words based on their probability of being used together in the same document. 
The PMI score for a pair of words $(w_1,w_2)$ is given by:
\begin{equation}
PMI(w_1,w_2)=\log{\frac{P(w_1 w_2)}{P(w_1)P(w_2)}}
\label{eq:pmi}
\end{equation}
The probabilities of the words and their co-occurrence $P(w_1)$, $P(w_2)$, $P(w_1 w_2)$ are obtained from our Vox corpus.
To quantify the topic coherence of each cluster of documents, we use $\widehat{PMI}$, the median PMI score between its 15 most common words (changing the number of words considered gives similar results). 

Finally, to obtain an aggregate topic coherence score for a set of clusters, we take the weighted average of the $\widehat{PMI}$ cluster scores.
The PMI score has been shown as best performing~\citep{pmi_coherence, pmi_coherence2} when compared to human interpretation of topics on different applications and corpora~\citep{pmi_coherence_lda, twitter_pmi_coherence}. 

\paragraph*{Comparing different partitions}\label{sec:nmi}
To compare how similar two partitions $T$ and $C$ of the same graph are, we use the normalised mutual information (NMI):
\begin{equation}
\label{eq:nmi}
NMI(T,C)=\frac{I(T,C)}{\sqrt[]{H(C)H(T)}}=\frac{\sum_{t \in T} \sum_{c \in C} p(t,c) \log{\left(\frac{p(t,c)}{p(t)p(c)}\right)}}{\sqrt[]{H(C)H(T)}}
\end{equation}
where $I(T,C)$ is the Mutual Information, and $H(T)$ and $H(C)$ are the entropies of the two cluster assignments. The score is bounded $0 \leq NMI \leq 1$, and an increased value of $NMI$ reflects higher similarity of the partitions. We will use the $NMI$ score to compare the partitions obtained by MS and other methods against the classification assigned by commercial services.

\section{Results}

\subsection{Multiscale clustering of the Vox corpus of news articles}
Figure~\ref{fig:MS} shows the results of applying Markov Stability to the similarity graph of articles constructed from the 9021 news articles of the Vox corpus published during 2016. MS sweeps across all scales (i.e., across Markov times) and reveals that the news corpus presents a high level of structure in its content, as can be seen by the existence of strong modular substructures at different levels of resolution (i.e., long plateaux of $VI(t,t')$  at a variety of Markov times with corresponding dips in $VI(t)$). This multiscale structure is indicative of the presence of groups of similar documents at different levels of granularity. 
To illustrate our results, we concentrate on four chosen resolution levels of increasing coarseness with 28, 15, 9, and 5 clusters of news articles.

The multi-level Sankey diagramme (Fig.~\ref{fig:MS}, bottom) shows that the clusters of similar documents present a quasi-hierarchical structure across scales, i.e., finer groups of similar articles integrate with other finer groups into coarser groups in a consistent manner. It is important to remark that this result emerges \emph{directly} from the structure of the data, as MS does not impose the existence of such a hierarchy in the multiscale clusterings. This organisation reflects the fact that the news clusters display a relationship based on content at different levels of granularity, as shown below.

\subsection{Describing the Topic Clusters}
\label{sec:description}
\begin{figure*}[h!]
\includegraphics[width=.99\linewidth,angle=0]{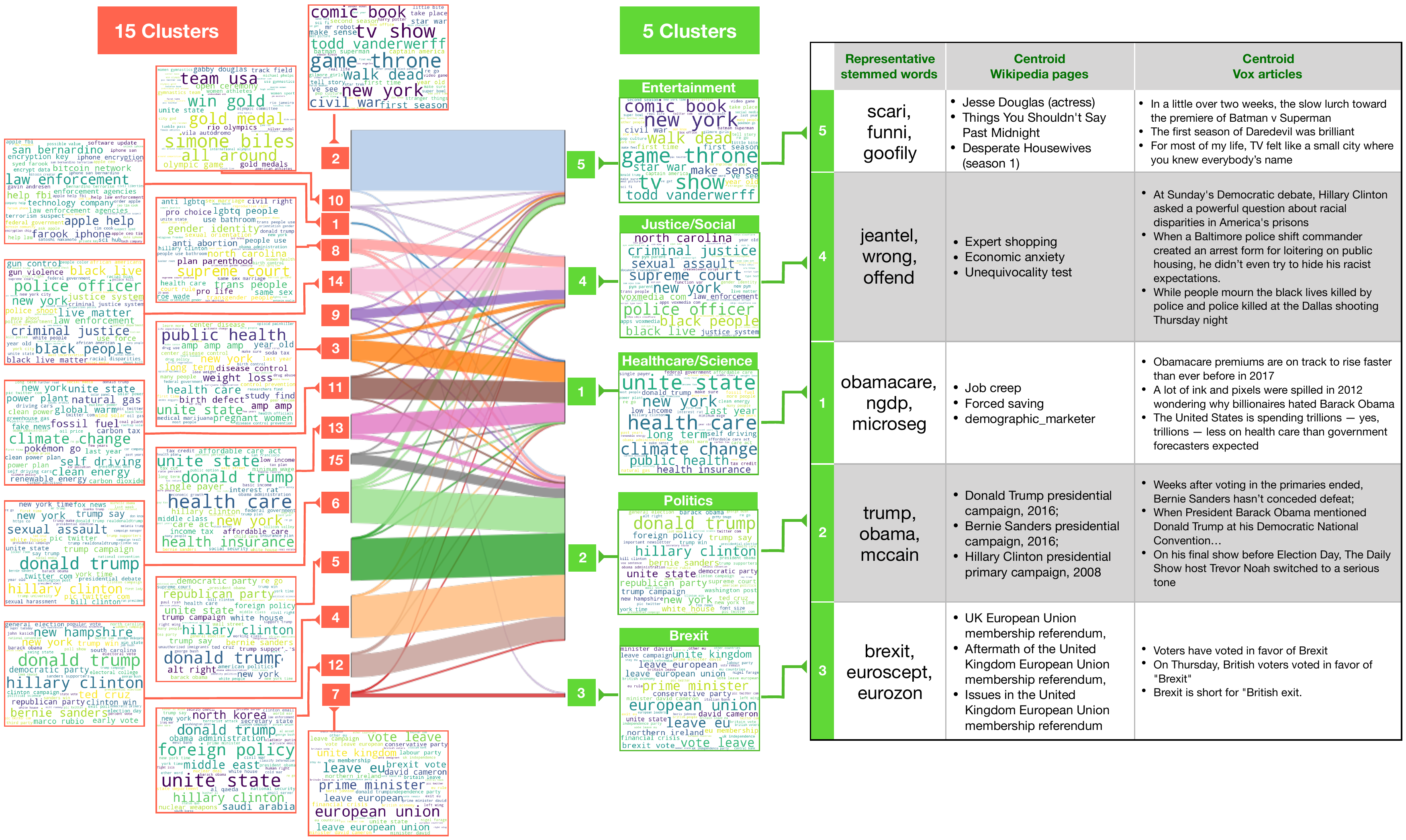}
\caption{
The Sankey diagramme illustrates the consistency between the finer partition of 15 clusters and the coarser partition of 5 clusters. We also show the corresponding word clouds of both levels, which highlight the specificity of the topics found in the text and its consistent integration into 5 broader areas (`Entertainment' 'Politics' 'Healthcare' `Social issues' and `Brexit'). Hence the multi-resolution coarsening in the content mirrors the multi-level, quasi-hierarchical community structure found in the document similarity graph. In addition to the word clouds, for the 5-cluster partition we also include the closest Words, closest Wikipedia page, and closest Vox article to the centroid of each of the clusters. Our vector representations allow us to find such mappings directly from the data, thus allowing a further use of the original data to aid the interpretability of the obtained clusters. 
}
\label{fig:15to5}
\end{figure*}

We illustrate the methodology with the analysis of selected levels of granularity. Figure~\ref{fig:15to5} illustrates the relationship between two of the scales in the MS partitioning (15 and 5 clusters) using a Sankey diagramme and word clouds that provide insight as to the content of the different clusters. 

As shown above, both levels show a strong quasi-hierarchical structure, with smaller `sub-topics' integrating to form larger `topics'.
This is particularly clear in the merger of several groups of documents (12, 4, 5, 6, 15 from the 15-way partition) to form the large 'Politics' group 2 in the 5-way partition. 
Similarly, the merger of the communities 3, 11, 13 from the 15-cluster partition (which correspond to environment, healthcare, and healthcare reform) give rise to the larger group 1 in the 5-way partition that contains science and technology issues (with health as a main component).
The group 4 in the 5-way clustering is the result of a merger of several groups related to justice, law enforcement and social issues, including a large component of the 'Black lives' campaign.
Other groups remain mostly unchanged between these two scales. For instance, the large group 2 of the 15-cluster partition dominates the 
even larger group 5 in the 5-way clustering corresponding to Entertainment, TV shows, movies and sport.
Interestingly, note that the relatively small group of articles referring to Brexit, remain in their own group (from group 7 in the 15-cluster partition to group 3 in the 5-cluster partition), as their content is highly distinctive and different to the mainly US-centric Vox news corpus.
This example highlights the sensitivity of the method to truly distinct topics in the corpus.

The fact that our method is based on vector embeddings for the documents has additional advantages. Not only does each document have a vector representations but each cluster can also be represented as a vector, i.e., the \textit{average vector sum} of all document vectors within the cluster. Once this cluster vector has been obtained, we can use our embedding model to assign to each cluster \textit{representative} words and documents from both the training set (Wikipedia in this case) and the dataset (Vox corpus). To assign the representative words and documents, we exploit the vector space representation, which allows us to define a distance metric, so that
representative instances are chosen to be geometrically close to the cluster centroid. In Figure~\ref{fig:15to5} we illustrate this procedure applied to the 5-way partition, where we show the three closest words (stemmed) from the Word2vec model vocabulary, the three closest Wikipedia pages (title reported), and the three Vox articles (first sentences reported) to the centroid of each of the five MS clusters.

The results of the 28-cluster partition also show the same quasi-hierarchical content structure, as seen in the Sankey diagramme of Fig.~\ref{fig:MS} and confirmed with the word clouds presented in Fig.~\ref{fig:Calais_Google} and the temporal profiles of some of the clusters in Fig.~\ref{fig:timelines}. Some of these features are discussed in more detail below.

\subsection{Evaluating the quality of the topic clusters}
\label{sec:quality}

Broad corpora, such as news, pose difficulties in the assignment of a `ground truth' for classification purposes. Traditionally, the categorisation of news is based non-exclusive labels denoting generic themes that cut across broad areas (e.g., 'News', 'Entertainment', 'Human interest', etc) and complemented by more specific categories (e.g., 'Weather', 'Law', 'Technology', etc). This approach usually leads to unbalanced groupings. In particular, the Vox news corpus analysed here does not contain a viable categorisation; a hand-coded 'Category' column in the dataset is highly unbalanced with almost half of the articles belonging to the category 'Latest'. 

Our analysis in Section~\ref{sec:description} reveals that the content-based MS clusters reflect consistent topics with different levels of detail. To evaluate the relevance of the clusters, and in the absence of an external ground truth, we follow two complementary routes.

\subsubsection{\textbf{Intrinsic Topic Coherence: Comparison Against Common Clustering Methods}}

We compute the median PMI score $\widehat{PMI}$~\eqref{eq:pmi} of the MS partitions as a measure of intrinsic topic coherence.

Furthermore, we compared the topic coherence of our MS results against clusterings produced with three widely-used, standard methods:
hierarchical clustering with Ward linkage~\citep{ward} with both Bag of Words TF-iDF vectors (Ward-BoW) and our own Doc2vec vectors (Ward-D2V)\footnote{BoW vectors generated using the default parameters of Scikit-Learn's~\citep{scikit-learn} TfidfVectorizer class. Ward clustering applied through the same module's AgglomerativeClustering class using 'wards' for the linkage parameter.};
and LDA probabilistic topic models obtained by using the Gensim~\citep{gensim} module (with 5 passes) to train over the Vox news dataset.

Table~\ref{table:tc_comm_comp} shows the topic coherence scores  different for each of the methods for different resolutions . (Note that for all methods, the value of $\widehat{PMI}$ increases for finer clusters, due to the dictionary containing more similar words in smaller groups.) 
MS and Ward-BoW clusters consistently provide higher topic coherence scores across resolutions.

\begin{table}[h]
\begin{tabular}{|l|l|l|l|l|}
\hline
\multicolumn{1}{|c|}{\textbf{$\widehat{PMI}$ scores}} & \multicolumn{1}{c|}{\textbf{28}} & \multicolumn{1}{c|}{\textbf{15}} & \multicolumn{1}{c|}{\textbf{9}} & \multicolumn{1}{c|}{\textbf{5}} \\ \hline
\textbf{MS D2V} & 0.23 & 0.257 & 0.136 & 0.127 \\ \hline
\textbf{Ward BoW} & 0.241 & 0.209 & 0.146 & 0.138 \\ \hline
\textbf{Ward D2V} & 0.212 & 0.156 & 0.1 & 0.071 \\ \hline
\textbf{LDA} & 0.144 & 0.145 & 0.109 & 0.124 \\ \hline
\end{tabular}
\caption{Median Topic Coherence scores ($\widehat{PMI}$ based on the top 15 words) for different clustering methods (rows) at different levels of clustering resolution (number of clusters, columns).}
\label{table:tc_comm_comp}
\end{table}

\subsubsection{\textbf{Comparisons against Commercial Taxonomy and Classification Services}}

\begin{figure*}[h!]
\includegraphics[width=.9\linewidth,angle=0]{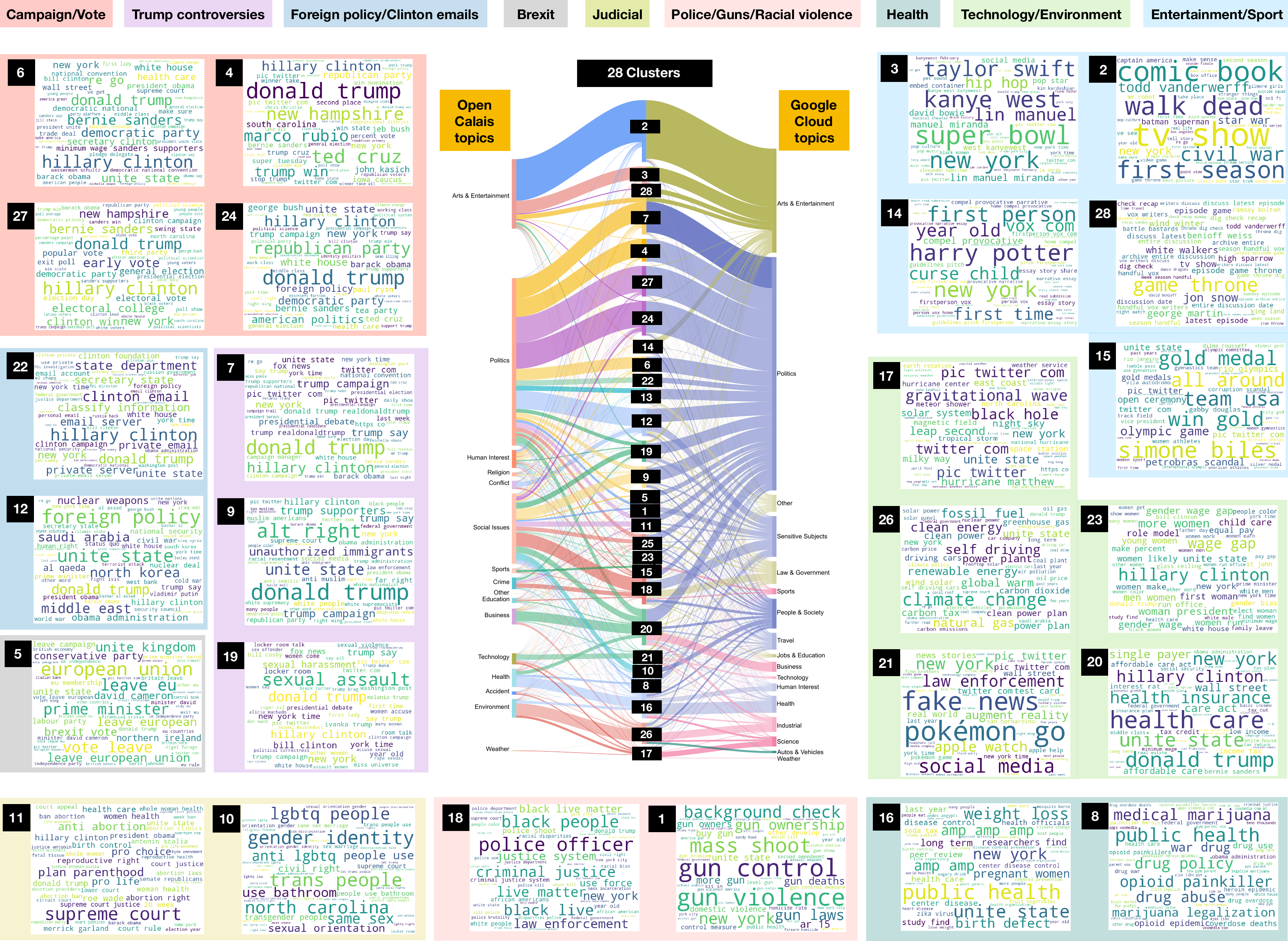}
\caption{Sankey diagramme comparing the 28-way MS clustering against the classification provided by external taxonomy services Open Calais and Google Cloud. The external services produce unbalanced classifications, with large generic categories such as 'News' and 'Entertainment'. Our partition is consistent across both services, and provides additional detail for the categories, as indicated by the word clouds and the coloured groupings (see legend) derived from our multilayer Sankey diagramme in Figure~\ref{fig:MS}.
}
\label{fig:Calais_Google}
\end{figure*}

The MS clusters of documents are purely content-driven with no external labeling or categorisation. Yet the clusters capture topical information and we expect that, in many cases, the content-driven groupings will be well aligned with standard categories used in news classifcation. 

To explore this issue, we turned to comparisons of the clusters with external `news ontologies'. 
We used two external, well-used commercial sources for news categorisation: the Natural Language API of Google Cloud Platform~\footnote{Google has opened its text analysis capabilities as a commercial service under the name of Google Cloud Natural Language. More information is available on \url{https://cloud.google.com/natural-language/}} and the Open Calais API of Thomson Reuters~\footnote{Thomson Reuters has an API service, accessible from \url{http://www.opencalais.com/}}. 
We used Python Requests Library and standard demo accounts on both of these services for comparison purposes.
For Google Cloud, 
we used only the top category 
and we manually merged Google Cloud topics with high similarity (e.g., 'Weather' and 'News/Weather'; 'Finance' and 'News/Finance', etc).  
Open Calais specialises in news and research topics 
and uses the news classifications according to the International Press Telecommunications Council (IPTC)'s taxonomy~\footnote{Full list of Open Calais supported IPTC topics at \url{http://www.opencalais.com/wp-content/uploads/folder/ThomsonReutersOpenCalaisAPIUserGuideR11_7.pdf}}. 
In our case, Open Calais was not able to classify a significant fraction of the Vox corpus. We stored those articles as 'Unclassified' according to Calais. 

We utilised both Google Cloud and Open Calais to classify our Vox news corpus into categories. Both classification services have resulted to 17 categories, including two ambiguous ones: 'Unclassified' and 'Others'. Hence the most relevant level of resolution for the MS clustering to compare against these services is the 15-way partition.
Interestingly, the topic coherence for the commercial services is higher than that achieved by LDA but lower than the one achieved by MS with a similar number of clusters: $\widehat{PMI}$(LDA-15)=0.144 < $\widehat{PMI}$(Open Calais)=0.157 < $\widehat{PMI}$(Open Google Cloud)=0.204 < $\widehat{PMI}$(MS-15) =0.257 (see Table~\ref{table:tc_comm_comp}).

To compare our clusters to the classifications obtained by the commercial taxonomy services, we obtained the Normalized Mutual Information (NMI) scores~\eqref{eq:nmi} between MS and other methods against Google Cloud and Open Calais classifications(Table~\ref{table:NMI_scores}). Our results show that MS has the highest similarity to both commercial services, thus emphasising the relevance of the clusters as compared to classifications that employ user-based knowledge.  
\begin{table}[h]
\begin{tabular}{|l|c|c|c|c|}
\hline
\multicolumn{1}{|c|}{\textbf{NMI Scores}} & \textbf{MS D2V} & \textbf{LDA} & \textbf{Ward D2V} & \textbf{Ward BoW} \\ \hline
\textbf{Open Calais} & 0.303 & 0.263 & 0.270 & 0.202 \\ \hline
\textbf{Google Cloud} & 0.351 & 0.330 & 0.322 & 0.266 \\ \hline
\end{tabular} 

\caption{The NMI scores between the four clustering methods (15 clusters) against the two commercial taxonomy services show that the MS clustering is closest to the output of both commercial classifications.}
\label{table:NMI_scores}
\end{table}

\begin{figure*}[h!]
\includegraphics[width=.8\linewidth,angle=0]{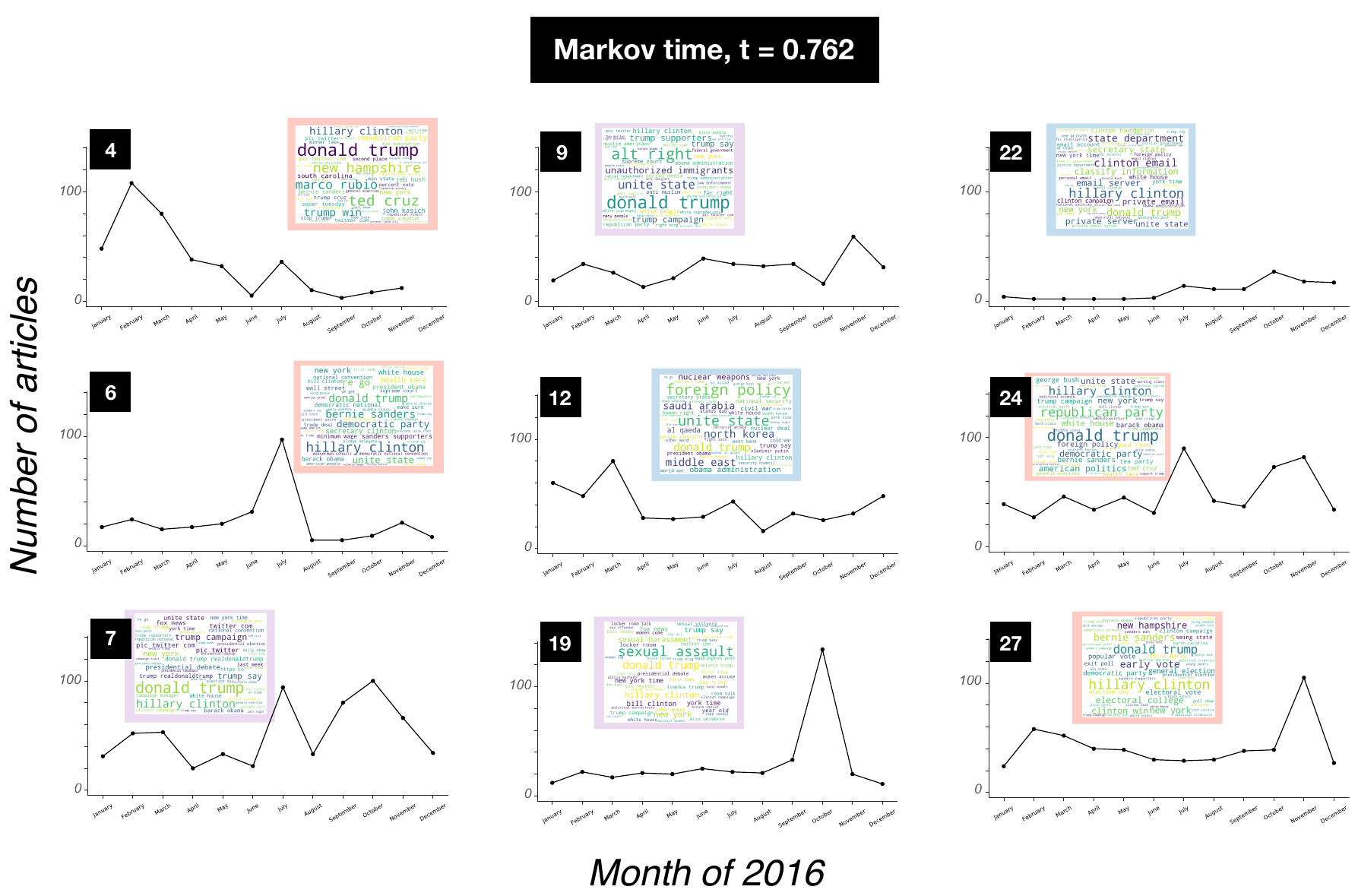}
\caption{The time profile of number of articles per month published during 2016 for 9 of the communities (in the 28-way partition) which are associated with the US election (see Fig.~\ref{fig:Calais_Google}). The temporal patterns show that the communities are related to events associated with the US electoral cycle (e.g., primaries, nomination, convention, election) as well as particular events that took place during the 2016 US campaign related to Trump and Clinton controversies.
}
\label{fig:timelines}
\end{figure*}

This correspondence can also be seen in Figure~\ref{fig:Calais_Google}, where we present the Sankey diagramme that compares of our finer 28-cluster partition with both Google Cloud Natural Language API and Open Calais API. As expected,  several of our communities fall consistently within the broad categories of 'News' and 'Entertainment' across both Google Cloud Natural Language and Open Calais. Yet the analysis also reveals the finer nuances and groupings that the content-driven categories can bring to the fore, as shown by the word clouds of 26 communities (of the 28, since 2 are dominated by wrapper artifacts embedded in the text). Using the multi-layer Sankey diagramme of our MS analysis in Fig.~\ref{fig:MS} and the word clouds we establish groupings of documents that fall within a branch of related content, thus allowing a deeper understanding of the different topics appearing in the corpus. In Fig~\ref{fig:Calais_Google}, we have coloured these branches and the legend indicates their general topic. For instance, there are three groupings related to US politics and the US election: 'Campaign/Vote', 'Trump', and 'Foreign Secretary/Clinton emails' involving 9 communities of the 28 groupings. A more detailed investigation of these clusters shows that they are thematically related to Republican and Democrat topics, and also to temporal events that took place during 2016. 
To see this explicitly, we have plotted in Figure~\ref{fig:timelines} the time line for each of those 9 communities, which show distinct temporal profiles. Several of these clusters are localised in time (e.g., clusters 4, 6, 19, 27) while others are constant throughout the year. The groupings can be directly related to different aspects of the US electoral cycle (primaries, nominations, national conventions, campaign) as well as particular events particular to the 2016 US campaign, such as the different Trump and Clinton controversies.

\section{conclusions}
We have introduced here a methodology for extracting multi-scale topic clusters from a corpus of text documents using no \text{a priori} labelling, in an unsupervised manner.  
Our application on a corpus of 9201 Vox media news articles shows consistency of topic clusters across different resolutions. The use of vector embeddings allows us to exploit the capabilities of novel text tools to aid with enhanced interpretation. 
Future work will be aimed at comparative analyses of multiple news outlets from different geographies and/or different political tendencies in order to enable quantification of the similarity and difference in topics each media outlet presents based on content.

\bibliographystyle{ACM-Reference-Format}
\bibliography{appnetsci}

\end{document}